\documentclass{article}

\usepackage{PRIMEarxiv}

\usepackage[utf8]{inputenc} 
\usepackage[T1]{fontenc}    
\usepackage{hyperref}       
\usepackage{url}            
\usepackage{booktabs}       
\usepackage{amsfonts}       
\usepackage{nicefrac}       
\usepackage{microtype}      
\usepackage{lipsum}
\usepackage{fancyhdr}       
\usepackage{graphicx}       
\graphicspath{{media/}}     

\usepackage{booktabs}
\usepackage{multirow}
\usepackage{xcolor}
\usepackage{amsbsy}
\usepackage{array}
\usepackage{amsmath}
\usepackage{amssymb}
\usepackage{algorithm}
\usepackage{algpseudocode}

\title{StochCA: A Novel Approach for Exploiting Pretrained Models with Cross-Attention
}

\author{
  Seungwon Seo\thanks{Equal contribution.} ~~~~~~~~~~~~~ Suho Lee\textsuperscript{*} ~~~~~~~~~~ Sangheum Hwang\thanks{Corresponding author.} \\
  \\
  Department of Data Science, \\
  Seoul National University of Science and Technology, \\
  Seoul, Republic of Korea \\
  \\
  \texttt{\{swseo, swlee\}@ds.seoultech.ac.kr}, \texttt{shwang@seoultech.ac.kr} \\
}

\begin{document}
\maketitle

\begin{abstract}
Utilizing large-scale pretrained models is a well-known strategy to enhance performance on various target tasks. It is typically achieved through fine-tuning pretrained models on target tasks. However, na\"{\i}ve fine-tuning may not fully leverage knowledge embedded in pretrained models. In this study, we introduce a novel fine-tuning method, called stochastic cross-attention (StochCA), specific to Transformer architectures. This method modifies the Transformer's self-attention mechanism to selectively utilize knowledge from pretrained models during fine-tuning. Specifically, in each block, instead of self-attention, cross-attention is performed stochastically according to the predefined probability, where keys and values are extracted from the corresponding block of a pretrained model. By doing so, queries and channel-mixing multi-layer perceptron layers of a target model are fine-tuned to target tasks to learn how to effectively exploit rich representations of pretrained models. To verify the effectiveness of StochCA, extensive experiments are conducted on benchmarks in the areas of transfer learning and domain generalization, where the exploitation of pretrained models is critical. Our experimental results show the superiority of StochCA over state-of-the-art approaches in both areas. Furthermore, we demonstrate that StochCA is complementary to existing approaches, i.e., it can be combined with them to further improve performance. Our code is available at \url{https://github.com/daintlab/stochastic_cross_attention}
\end{abstract}


\section{Introduction}
\label{sec:Int}
Recently, notable developments have emerged in large-scale models pretrained on extensive collections of images~\cite{Zhai_2022_CVPR, yuan2021florence,dehghani2023scaling,singh2023effectiveness}, natural language~\cite{smith2022using, chowdhery2022palm,driess2023palm,touvron2023llama,achiam2023gpt}, or multi-modal datasets~\cite{radford2021learning, alayrac2022flamingo,srivastava2024omnivec,wang2023one}. Due to the generic representations learned from these large-scale datasets, utilizing pretrained models is a well-known strategy for achieving successful performance in a wide range of downstream tasks. Pretrained models can potentially learn diverse semantic information that appears across large amounts of data rather than focusing on domain-specific patterns, such as image style. As a result, models that are well-transferred can be robust against distribution shifts, as demonstrated in recent domain generalization studies~\cite{cha2022miro,li2023simple}. Since pretrained weights provide a strong starting point for optimization, fine-tuning has become the \textit{de facto} standard for transferring pretrained knowledge to target tasks. However, na\"{\i}ve fine-tuning, which involves end-to-end training to update all model parameters, may not always be the optimal approach. For instance, na\"{\i}ve fine-tuning often requires a substantial amount of training data for target tasks to achieve satisfactory performance~\cite{chen2019catastrophic}. It can also distort the pretrained features, leading to reduced generalization performance on out-of-distribution data~\cite{kumar2022finetuning}. 
Therefore, with the increasing trend towards large-scale pretraining, it is crucial to effectively leverage target-specific information from the vast amount of pretrained knowledge.

In recent years, large-scale pretrained models in computer vision have often been built on the Vision Transformer (ViT) architecture~\cite{radford2021learning,kirillov2023segment,liu2023grounding}. The ViT~\cite{dosovitskiy2021image} is an architecture directly inherited from the Transformer~\cite{vaswani2017attention} for natural language processing. It processes image patches as input tokens, which undergo linear embeddings. A learnable embedding vector of a special \texttt{[CLS]} token is added to form an input sequence. This input sequence is then processed through multiple Transformer encoder blocks, resulting in a sequence of feature vectors. For image classification tasks, the final feature vector corresponding to the \texttt{[CLS]} token is typically used for class prediction. Due to its impressive performance, ViT is gradually replacing CNN-based models in a wide range of computer vision tasks. 

The self-attention module, a key component of ViT, assigns weights to different elements of an input sequence by computing attention scores. This mechanism allows a model to capture long-range dependencies and contextual information, thereby enhancing its ability to understand the relationships between visual elements (i.e., image patches). This self-attention has been extended to cross-attention for specific purposes in several previous studies. For example, Chen et al.~\cite{Chen_2021_ICCV} introduced a dual-branch architecture to vary patch resolution and incorporated cross-attention between the \texttt{[CLS]} token and patch tokens from different branches, enabling multi-scale feature representation learning. Jang et al.~\cite{jang2022unifying} used cross-attention to focus on relevant image regions while processing textual information, facilitating a better understanding of visual context and improving performance on tasks involving the fusion of image and text modalities. However, to the best of our knowledge, the potential of cross-attention as a method for transferring pretrained knowledge has not been explored in the literature.

In this study, we introduce Stochastic Cross-Attention (StochCA), a novel fine-tuning method designed to effectively leverage pretrained knowledge within the ViT architecture, tailoring it to target tasks. During training on a target task, cross-attention is employed to selectively reference the pretrained useful knowledge obtained from large-scale datasets. The rationale for using cross-attention lies in viewing the pretrained model as a key-value store that contains rich and valuable feature representations. In this framework, a target model's query vectors are fine-tuned to effectively access this store and retrieve meaningful representations for a given target task. The target model then learns to mix channel information by fine-tuning position-wise multi-layer perceptron (MLP) layers using the retrieved features.
In our approach, a target model, initialized with the same weights as the pretrained model, is trained on a specific target task. The pretrained model remains fixed (frozen) during training. In each attention layer of the target model, either self-attention or cross-attention is performed for each forward pass, determined by a predefined probability. If a layer is set to perform cross-attention, its output is computed using queries from the target model along with keys and values from the corresponding block of the pretrained model. Throughout the training, the target model's query vectors and MLP layers are trained to be compatible with both its own key and value pairs and those of the pretrained model. This compatibility is crucial for producing accurate predictions. In essence, the queries and MLP layers are fine-tuned to extract and utilize valuable knowledge specific to the target task from the pretrained model. As a result, the target model is expected to selectively incorporate this pretrained knowledge to effectively address the target task.

To validate the effectiveness of the proposed StochCA, we conducted evaluations under two distinct experimental settings: transfer learning (TL) and domain generalization (DG), which can benefit from the effective use of pretrained features. In the TL setting, which directly aligns with our primary objective of leveraging pretrained knowledge for a target task, we measured the performance on the target task as an indicator of the method's effectiveness. The DG setting focuses on model performance in unseen domains that share semantic similarities but differ in input distribution, such as image style. This setting tests the model's ability to identify domain-agnostic patterns, thus assessing the robustness of the target model to domain shifts.
In both the TL and DG settings, the proposed StochCA has been compared with state-of-the-art methods using several popular benchmarks. Our experimental results demonstrate the superiority of StochCA in both scenarios. Additionally, the versatility of StochCA is highlighted by its compatibility with other existing methods, showing improved performance when integrated with existing state-of-the-art approaches. This highlights the complementary nature of StochCA.
Furthermore, our analysis of the cosine similarity between the query, key, and value vectors of the target model and the pretrained model provides evidence that StochCA selectively exploits useful knowledge from the pretrained model, reinforcing the effectiveness of our approach.

In summary, our contributions are:
\begin{itemize}
    \item We propose a novel fine-tuning method named StochCA, which utilizes cross-attention to selectively leverage useful information from large-scale pretrained models for a target task.
    \item The effectiveness of StochCA is demonstrated in two experimental protocols, transfer learning (TL) and domain generalization (DG), where the effective exploitation of pretrained models is critical.
    \item Extensive experiments reveal that StochCA achieves comparable, and in some cases superior, performance compared to state-of-the-art methods in each setting. Our experimental results also suggest that StochCA can effectively complement existing methods.
    \item It is shown that StochCA selectively exploits the valuable knowledge of pretrained models by analyzing the cosine similarity between query, key, and value vectors from a target model and those from a pretrained model.
\end{itemize}

\section{Related Work}
\label{sec:Rel}
\subsection{Transfer Learning}
Transfer learning focuses on the effective adaptation of pretrained models to target tasks. This research area is particularly important in scenarios with limited target data, where the challenge is to avoid overfitting or negative transfer. For example, L2-SP~\cite{xuhong2018explicit} aims to regularize the target model's weights so that these weights do not deviate from the weights of the pretrained model. Similarly, DELTA~\cite{li2019delta} preserves pretrained knowledge by regularizing features between models. REGSL~\cite{li2021improved} takes a layer-wise approach to weight regularization and incorporates a self-labeling scheme for better robustness under noisy labels. StochNorm~\cite{kou2020stochastic} regularizes the statistics in batch normalization layers by stochastically selecting either moving average statistics or mini-batch statistics to avoid overfitting during fine-tuning. BSS~\cite{chen2019catastrophic} reduces small singular values of output features based on the claim that direct regularization may even preserve the pretrained knowledge that has a negative effect on the target task. In addition, Co-tuning~\cite{you2020co} exploits not only the feature extractor of the pretrained model but also the classifier for full transfer to the target task. However, most of existing transfer learning methods have been validated in the context of convolutional neural networks (CNNs) rather than ViT. In contrast, our proposed method is specifically designed for the ViT architecture, responding to its growing popularity and distinct characteristics. 

\subsection{Domain Generalization}
Domain generalization focuses on overcoming the challenge of domain shift, aiming to develop models that perform well on unseen target domains by training on multiple source domains. Recent research suggests that preserving features learned from large-scale datasets can effectively generalize to these new domains. For example, Cha et al.~\cite{cha2022miro} demonstrate that successful domain generalization can be achieved by maximizing the mutual information between features from a large-scale pretrained model and a model being trained with source domains. Li et al.~\cite{li2023simple} propose an efficient approach that involves learning adapters to merge predictions from a pool consisting of multiple pretrained models with different domain generalization capabilities, without the need to fine-tune the original models. Furthermore, the application of ViT to domain generalization has shown promising results. Zheng et al.~\cite{zheng2022prompt} experimentally observe that traditional domain generalization methods do not work well for ViT, and propose DoPrompt which learns to generate adaptive domain prompts for input images. Sultana et al.~\cite{sultana2022self} introduce a self-distillation approach that exploits the monolithic structure of ViT to minimize overfitting to source domains.

\subsection{Cross-attention in Vision Transformer}
As ViT is widely used in the field of computer vision, cross-attention has been applied in various studies for specific purposes. Dai et al.~\cite{dai2022cadg} introduce cross-attention between pairs of images of different styles, aiming to discern semantic information accurately irrespective of stylistic variations. Kim et al.~\cite{Kim_2022_ACCV} enhance video interpolation by applying cross-attention mechanisms between the query of a preceding frame and the key, value of a subsequent frame within two adjacent frames. In addition, numerous studies~\cite{zhu2022dual,ilinykh-dobnik-2022-attention,lin2022cat,jang2022unifying} apply cross-attention to capture better local features or to integrate multimodal information. In contrast to these studies, we utilize cross-attention to selectively refer to the useful knowledge of a pretrained model with large-scale datasets when transferring it to the target task.

\section{Method}
\label{sec:Met}
In this paper, we propose Stochastic Cross-Attention (StochCA), which selectively refers to the pretrained knowledge, and demonstrate its effectiveness on two experimental settings: transfer learning and domain generalization. First, we clarify the problem statement of the two experimental settings in Section~\ref{sec:problem}. Since our proposed method is developed by modifying the self-attention modules in ViT, we then present an overview of the self-attention mechanism in Section~\ref{sec:self_attention}. Next, we describe the proposed method, StochCA, in detail in Section~\ref{sec:stochca}.

\subsection{Problem Statement}\label{sec:problem}
\subsubsection{Transfer Learning}
As we focus on the classification task, a network $f$ consists of a feature extractor $F$ and a classifier $C$. Given a model $f_0$ pretrained on a large-scale source dataset $\mathcal{D}_s=\{(x_i^s,y_i^s)\}^{N_s}_{i=1}$, the goal of transfer learning is to obtain a model $f_t$ that performs well on a target task by fine-tuning $f_0$ with a target dataset $\mathcal{D}_t=\{(x_i^t,y_i^t)\}^{N_t}_{i=1}$. In the context of transfer learning, $\mathcal{D}_s$ and $\mathcal{D}_t$ typically share a similar input space but differ in their category spaces. For example, in computer vision tasks, $D_s$ often represents large-scale datasets such as ImageNet~\cite{deng2009imagenet}, while $D_t$ refers to a specific visual classification dataset of interest~\cite{you2020co}. Given that the label spaces of $\mathcal{D}_s$ and $\mathcal{D}_t$ are different, the pretrained model $f_0$ cannot be directly applied to the target dataset $\mathcal{D}_t$. To address this, a task-specific module of $f_0$ (i.e., the classifier $C$) is replaced with a new classifier $C'$ that is randomly initialized and tailored to fit the label space of the target task. Then, the feature extractor $F$ of $f_0$ equipped with the target-specific classifier $C'$ is fine-tuned for the target task to obtain the target model $f^*$:
\begin{equation}
    f^* = \underset{F, C'}{\arg\min} \frac{1}{\mathcal{D}_t}\sum_{i=1}^{N_t}\ell(C'(F(x_i^t)),y_i^t),
\end{equation}
where $\ell(\cdot,\cdot)$ is a loss function such as cross-entropy.

\subsubsection{Domain Generalization}
In domain generalization, a dataset for a target task consists of multiple domains. Let $\mathcal{D}_s=\{\mathcal{D}_1, \mathcal{D}_2, ..., \mathcal{D}_n\}$ be source domains, where each $\mathcal{D}_k$ represents a domain with $\{(x_i^{s(k)},y_i^{s(k)})\}^{N_s^{(k)}}_{i=1}$. The goal of domain generalization is to train a model that performs well on an unseen target domain $\mathcal{D}_t=\{(x_i^{t},y_i^{t})\}^{N_t}_{i=1}$ by accessing only the source domains. Unlike transfer learning, both $D_s$ and $D_t$ share the same label spaces, but they differ in input distributions due to domain shifts such as variations in image style. During training in domain generalization, data from the specific target domain $D_t$ is not accessible: the model can only use data from the source domains $D_s$. Therefore, it is crucial for the model to learn domain-invariant representations that encapsulate the underlying knowledge shared across all source domains. Similar to transfer learning, a pretrained model $f_0$ on a large-scale dataset such as ImageNet~\cite{deng2009imagenet} is generally employed as a starting point. This pretrained model is then further trained on $D_s$ to adapt to domain generalization tasks. The objective of vanilla fine-tuning for domain generalization is to optimize model parameters to minimize the loss values across all source domains, which can be formulated as:
\begin{equation}
    f^* = \underset{F, C'}{\arg\min} \frac{1}{\mathcal{D}_s}\sum_{k=1}^n \sum_{i=1}^{N_s^{(k)}}\ell(C'(F(x_i^{s(k)})),y_i^{s(k)}),
\end{equation}
where $\ell(\cdot,\cdot)$ represents a loss function such as cross-entropy.

\subsection{Self-Attention of Vision Transformer}\label{sec:self_attention}
The self-attention (SA) module is a pivotal component in ViT, responsible for capturing long-range dependencies between image patches. This is achieved by assigning importance weights to different spatial positions (i.e., image patches) within an image, enabling the model to focus on relevant features during training. By leveraging self-attention, ViT can effectively model the relationships between image patches and has achieved state-of-the-art performance in various computer vision tasks. 

Consider $X\in \mathbb{R}^{n\times d}$ as an input sequence to the SA layer, where $n$ represents the number of tokens and $d$ is the hidden dimension. The query $Q\in \mathbb{R}^{n\times d_q}$, key $K\in \mathbb{R}^{n\times d_k}$, and value $V\in \mathbb{R}^{n\times d_v}$ are defined and processed as follows:
\begin{equation}\label{eq:qkv}
    Q=XW^Q, K=XW^K, V = XW^V
\end{equation}
\begin{equation}\label{eq:SA}
    SA(Q, K, V) = Softmax(QK^\top / \sqrt{d_K})V,
\end{equation}
where $W^Q \in \mathbb{R}^{d\times d_q}$, $W^K \in \mathbb{R}^{d\times d_k}$, and $W^V \in \mathbb{R}^{d\times d_v}$ are weight matrices for calculating $Q$, $K$, and $V$, respectively. For simplification, this explanation considers single-head self-attention where $d_q = d_k = d_v = d$. The self-attention module thus learns to focus within the input sequence itself by generating queries, keys, and values from the same input.

\begin{figure}[t]
    \centering
    \includegraphics[width=0.65\textwidth]{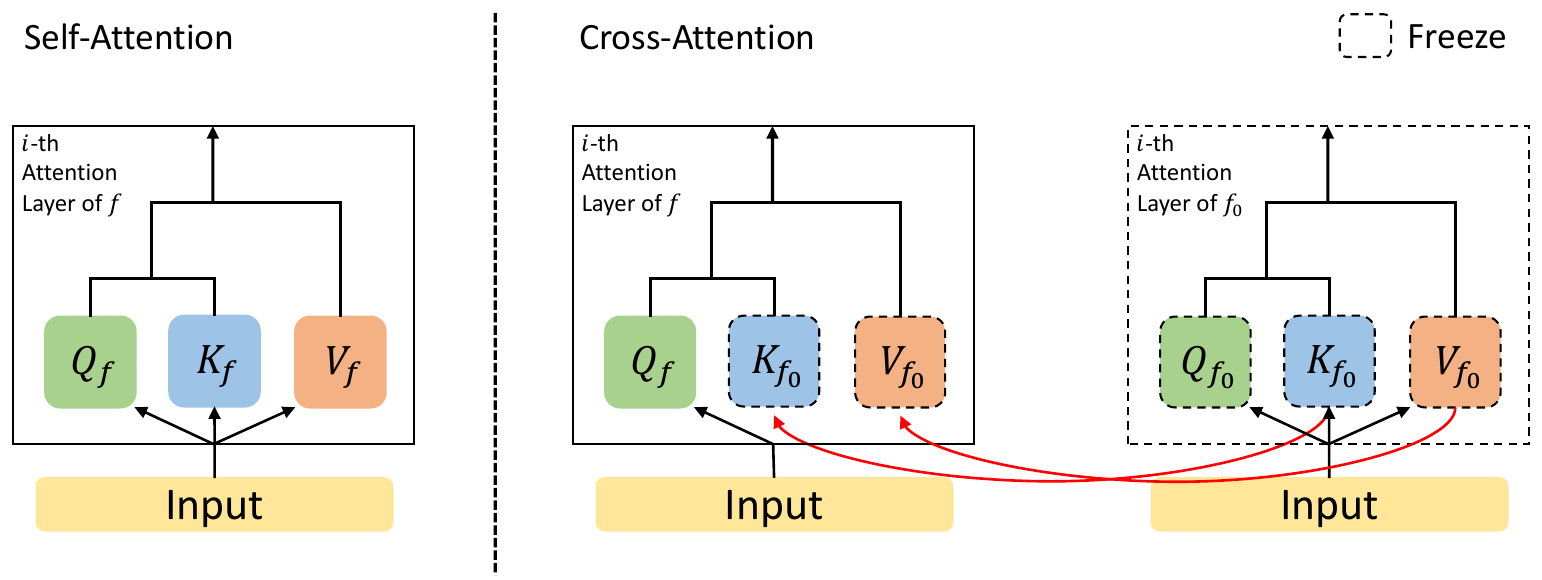}
    \caption{
    Illustration of the attention mechanisms: (Left) The process of self-attention, where a model focuses on different parts of a single input sequence to compute its representations. (Right) The process of cross-attention, where a model attends to a different sequence (e.g., from a pretrained model) while processing an input sequence, facilitating an interaction between the two.}
    \label{fig:saca}
\end{figure}

\subsection{Stochastic Cross-Attention}\label{sec:stochca}
A characteristic of the self-attention module is that query, key, and value are derived from the same input sequence $X$. In contrast, cross-attention (CA) has been employed in various studies to model conditional information by calculating the query, key, and value from different inputs, making it suitable for specific applications~\cite{dai2022cadg,Kim_2022_ACCV,zhu2022dual,ilinykh-dobnik-2022-attention,lin2022cat,jang2022unifying}. In this work, we utilize cross-attention to selectively access relevant knowledge from a large-scale pretrained model during training on a given target task. 

Let $f$ denote a target model being trained on a given target dataset, and $f_0$ be a pretrained model that the target model refers to. We assume that both the target and the pretrained model share the same architecture. The cross-attention mechanism for referencing features of the pretrained model at the $l$-th attention layer is computed as: 
\begin{equation}\label{eq:CA}
    CA(Q_f^l, K_{f_0}^{l}, V_{f_0}^{l}) = Softmax(Q_f^{l} K_{f_0}^{l\top} / \sqrt{d_K})V_{f_0}^{l},
\end{equation}
where $Q_f^l$ is the query obtained from the $l$-th attention layer of the target model $f$, and $K_{f_0}^{l}$ and $V_{f_0}^{l}$ are the key and value from the $l$-th attention layer of the pretrained model $f_0$, respectively. Through this cross-attention, the query of the target model $f$ learns to efficiently extract useful information relevant to the target task from the key and value of the pretrained model $f_0$. Consequently, it enables the target model $f$ to selectively leverage the knowledge embedded in the pretrained model $f_0$. Building upon these representations, the MLP layers in the target model are fine-tuned, which focuses on learning channel-mixing strategies that are particularly relevant to the target task.
Figure~\ref{fig:saca} illustrates the processes of self-attention and cross-attention with a pretrained model, providing a visual representation of these concepts.

\begin{figure}[!t]
    \centering
    \hspace{1mm}
    \includegraphics[width=0.88\textwidth]{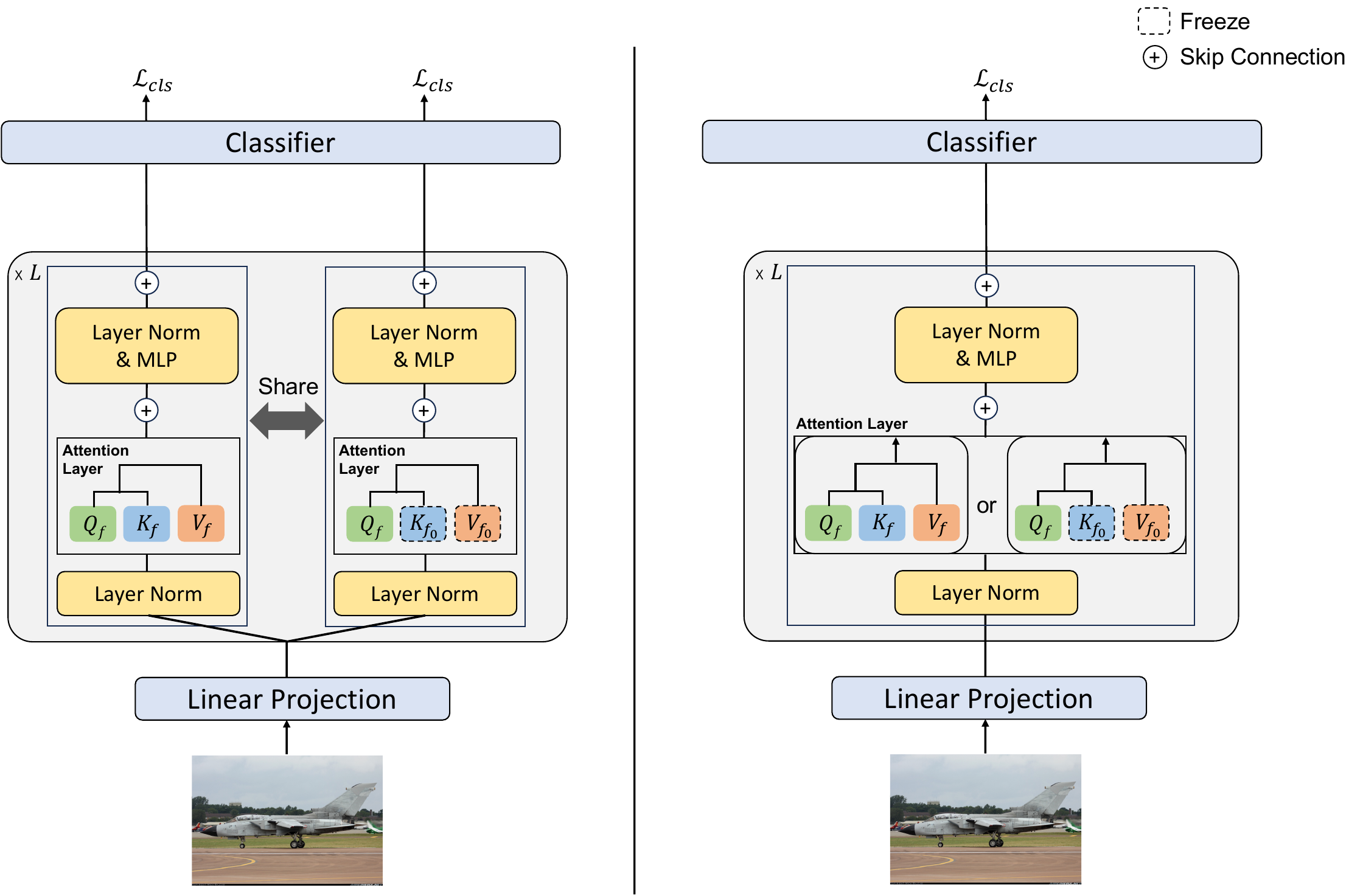}
    \caption{Two approaches to combining self-attention and cross-attention: (Left) A simple framework that integrates both self-attention and cross-attention in a straightforward manner. (Right) The StochCA Framework that incorporates stochasticity in the combination of self-attention and cross-attention.}
    \label{fig:simple}
\end{figure}

However, if all self-attention layers in the target model $f$ are replaced by cross-attention, learning for the target task may be insufficient due to over-reliance on the pretrained model $f_0$. To balance this, an appropriate utilization of both self-attention and cross-attention is required. One straightforward method, as depicted in Figure~\ref{fig:simple} (left), involves performing both self-attention and cross-attention in every attention layer, with the final prediction being an average of the outputs from both paths. However, this approach necessitates the use of the pretrained model during inference for the cross-attention path, thus increasing memory and computational requirements. Additionally, it requires dual propagation for each image (one each for self-attention and cross-attention paths), leading to higher computational costs during training.

To enhance computational efficiency, we propose the block-wise stochastic cross-attention (StochCA) method, which selectively references the representations of the pretrained model. In StochCA, each attention layer of the target model $f$ is assigned a probability $p$ for performing cross-attention. Based on this probability, the model stochastically selects either self-attention or cross-attention during each training step. Specifically, the output of the $l$-th attention layer is computed as follows: 
\begin{equation}
    h_l = (1-\beta)SA(Q_f^l, K_f^l, V_f^l) + \beta CA(Q_f^l, K_{f_0}^{l}, V_{f_0}^{l}),
\end{equation}
where $\beta$ is a random variable of \textit{Bernoulli} distribution with probability $p$.  Compared to the vanilla ViT, cross-attention is performed instead of self-attention stochastically in each attention layer. During inference, the pretrained model $f_0$ is no longer needed and the output computed via only self-attention is used for the final prediction (i.e., $p=0$), thus avoiding additional computational demands. The hyperparameter $p$ regulates the target model's dependency on $f_0$. The overall framework of StochCA is illustrated in Figure~\ref{fig:simple} (right) and described in detail in Algorithm~\ref{StochCA_trn}.

\begin{figure}[!t]
\centering
\begin{minipage}{0.8\linewidth}
    \begin{algorithm}[H]
        \caption{Training procedure of Stochastic Cross-Attention}
        \label{StochCA_trn}
        \begin{algorithmic}
            \Require{pretrained model $f_0$, target model $f$, cross-attention probability $p$, number of iterations $N$, target training dataset $\mathcal{D}_{tr}$}
            \\
            \For {$iteration=1,2,...,N$}
            \State{Sample a data ($x_i, y_i$) from the training set $\mathcal{D}_{tr}$}
            \State{Extract $\{K^l_{f_0}, V^l_{f_0}\}_{l=1}^L$ for ($x_i, y_i$) from the pretrained model $f_0$}
            \State{$h_0 = E_f(x_i)$} \Comment{Image patch embeddings}
            \For {$l=1,2,...,L$}
            \If{$\beta~\sim Bernoulli(p) = 1$}
            \State{Generate $Q_f^l$ from $h_{l-1}$ using Eq.~\ref{eq:CA}}
            \State{$h_{l}=CA(Q_f^l,K_{f_0}^l,V_{f_0}^l)$} \Comment{Perform cross-attention} 
            \Else
            \State{Generate $Q_f^l,K_f^l,V_f^l$ from $h_{l-1}$ using Eq.~\ref{eq:SA}}
            \State{$h_{l}=SA(Q_f^l,K_f^l,V_f^l)$} \Comment{Perform self-attention}
            \EndIf        
            \EndFor
            \State{Update $f$ by minimizing cross-entropy loss $l(C_f(h_L), y_i)$}
            \EndFor
            \\
            \Ensure{Trained target model $f$}
        \end{algorithmic}
    \end{algorithm}
\end{minipage}
\end{figure}

\section{Experiments}
\label{sec:Exp}
\subsection{Dataset}
To assess the effectiveness of our proposed method, we utilize a total of eight image classification benchmarks commonly used in transfer learning and domain generalization. 

For transfer learning, our method is evaluated on the following four datasets:
\begin{itemize}
    \item Caltech-UCSD Birds-200-2011 (CUB)~\cite{WahCUB_200_2011}: A bird species classification dataset comprising 11,788 images across 200 bird species.
    \item Stanford Cars (Car)~\cite{krause20133d}: A dataset including 16,185 images of 196 car classes, featuring diverse car viewpoints, occlusions, and background variations.
    \item FGVC-Aircraft (Aircraft)~\cite{maji2013fine}: An aircraft classification dataset which contains 10,200 images of 100 different aircraft models from various viewpoints.
    \item Stanford Dog (Dog)~\cite{khosla2011novel}: A dog breed classification dataset containing 20,580 images of 120 different dog breeds.
\end{itemize}

For domain generalization, we evaluate our method on the following four datasets:
\begin{itemize}
    \item PACS~\cite{li2017deeper}: A dataset comprising four domains (Art, Cartoon, Photo, and Sketch) with 9,991 images in 7 classes.
    \item VLCS~\cite{fang2013unbiased}: A dataset consisting of four photographic domains (Caltech101, LabelMe, SUN09, and VOC2007) with 10,729 images in 5 classes.
    \item OfficeHome~\cite{venkateswara2017deep}: A dataset including four domains (Art, Clipart, Product, and Real) with 15,500 images in 65 classes easily found in office and home environments. 
    \item DomainNet~\cite{peng2019moment}: A large-scale visual recognition dataset for domain generalization that encompasses six domains (Clipart, Infograph, Painting, Quickdraw, Real, and Sketch), containing 586,575 images across 345 classes.
\end{itemize}

\begin{figure}[t]
    \centering
    \includegraphics[width=0.85\textwidth]{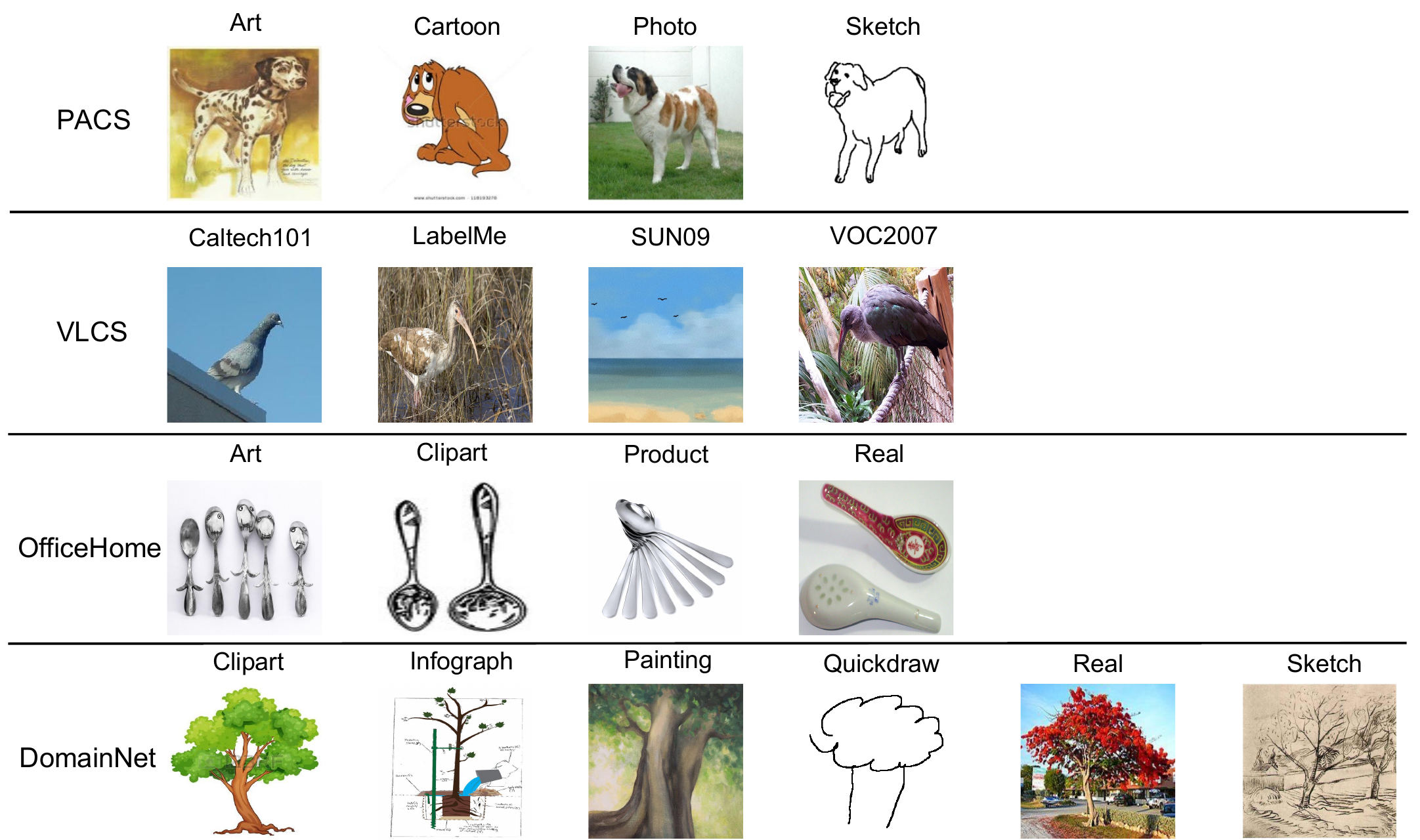}
    \caption{Samples of domain generalization benchmarks.}
    \label{fig:exam}
\end{figure}

Figure~\ref{fig:exam} illustrates examples from different domains for each dataset, showcasing the same class in varied styles.

\subsection{Comparison Methods}
For transfer learning, our proposed method is benchmarked against four established transfer learning algorithms: FT (i.e., vanilla fine-tuning), L2-SP~\cite{xuhong2018explicit}, BSS~\cite{chen2019catastrophic}, and Co-tuning~\cite{you2020co}. L2-SP~\cite{xuhong2018explicit} regularizes the weights of the fine-tuned model to prevent significant deviation from the pretrained model’s weights. BSS~\cite{chen2019catastrophic} minimizes small singular values of feature maps, thereby reducing the risk of negative transfer. Co-tuning~\cite{you2020co} focuses on fully leveraging a pretrained model by learning the label relationship between the source and target tasks. 

For domain generalization, popular DG algorithms in DomainBed~\cite{gulrajanisearch2021in} are employed as comparison targets. These include ERM~\cite{gulrajanisearch2021in}, IRM~\cite{arjovsky2019invariant}, CDANN~\cite{li2018deep}, DANN~\cite{ganin2016domain}, GDRO~\cite{Sagawa2020Distributionally}, CORAL~\cite{sun2016deep}, MLDG~\cite{li2018learning}, Mixup~\cite{wang2020heterogeneous}, and MMD~\cite{li2018domain}. Additionally, we compare our method with DoPrompt~\cite{zheng2022prompt}, which enhances performance by learning domain-specific prompts in the ViT structure, and MIRO~\cite{cha2022miro}, an approach that leverages pretrained knowledge by maximizing mutual information. 

\subsection{Implementation Details}
In our experiments, we utilize ViT-base/16 pretrained on ImageNet~\cite{deng2009imagenet} and the AdamW optimizer~\cite{loshchilov2018decoupled} as our default settings. In the case of transfer learning, models are trained for 10,000 steps with a batch size of 16. Following~\cite{zhu2022dual}, the base learning rate is set to ${5e^{-4}}/{512} \times \texttt{batchsize}$ with 100 steps of warm-up, followed by cosine scheduling for decay. We apply a $550\times550$ resize, a $448\times448$ random crop, and a random horizontal flip for augmentation during training. To ensure a fair comparison among different methods, we search for algorithm-specific hyperparameters (e.g., the coefficient $\beta$ in L2-SP) for each method based on validation accuracy. A subset of $10\%$ of the training dataset is used for hyperparameter tuning, except for the Aircraft dataset, which has an official validation set. After the hyperparameter selection, all models are re-trained on the combined training dataset (i.e., including both the training and validation datasets) with the selected hyperparameters. The final performance evaluation is then measured on the test dataset. For our proposed StochCA method, the cross-attention probability $p$ is explored over the set $\{0.1, 0.3, 0.5, 0.7\}$. The optimal value of $p$ is determined to be $0.5$ for the Dog dataset and $0.1$ for the remaining datasets. 

For domain generalization experiments, we follow the experimental settings of \cite{gulrajanisearch2021in}. In this setup, each domain is split into training and validation subsets. The model is trained using one domain as the target and the remaining domains as the source. Training is performed on all training subsets of the source domains, and hyperparameters are selected based on performance across all validation subsets of these source domains. The model's test performance is then evaluated on the unseen target domain. This procedure is repeated for each possible combination of source and target domains, and the average test performances are calculated to determine the final performance for each benchmark. Hyperparameters for each algorithm are divided into two categories: algorithm-agnostic and algorithm-specific. Initially, algorithm-specific hyperparameters are tuned following the same approach as in transfer learning, while keeping algorithm-agnostic hyperparameters at their default values. The default algorithm-agnostic hyperparameters are a learning rate, weight decay, and a dropout rate for the last layer, which are set to $1e^{-5}$, $1e^{-2}$, and $0$, respectively. For PACS, VLCS, and OfficeHome, hyperparameters are determined based on validation accuracy evaluated every 200 steps. For the larger-scale DomainNet, validation accuracy is evaluated every 1,000 steps for hyperparameter selection. Once algorithm-specific hyperparameters are set, we proceed to fine-tune algorithm-agnostic hyperparameters. The learning rate, weight decay, and the dropout rate for the last layer are searched within the ranges of $\{5e^{-5}, 1e^{-5}, 5e^{-6}\}$, $\{0, 1e^{-2}\}$, and $\{0, 0.1\}$, respectively. All other configurations remain consistent with \cite{gulrajanisearch2021in}. For the StochCA method, the cross-attention probability $p$ is selected as $0.1$ for PACS and DomainNet and $0.3$ for OfficeHome and VLCS. We report the average accuracy (\%) for three random seeds in all experimental results.

\begin{table}[!ht]
    \begin{center}
    \resizebox{0.8\textwidth}{!}{%
    \begin{tabular}{llcccc}
        \toprule
        \multirow{2}{*}{Dataset} & \multirow{2}{*}{Method} & \multicolumn{4}{c}{Sampling Rates} \\
        \cmidrule{3-6}
        & & $15\%$ & $30\%$ & $50\%$ & $100\%$ \\
        \midrule 
        \multirow{5}{*}{CUB} & FT & $56.09_{\pm 1.12}$ & $75.39_{\pm 1.02}$ & $81.60_{\pm 0.47}$ & $86.85_{\pm 0.42}$ \\
        & L2-SP & $55.64_{\pm 1.16}$ & $75.20_{\pm 1.06}$ & $81.50_{\pm 0.48}$ & $86.82_{\pm 0.46}$ \\
        & BSS & $55.60_{\pm 1.35}$ & $75.37_{\pm 1.13}$ & $81.57_{\pm 0.43}$ & $86.72_{\pm 0.28}$ \\
        & Co-tuning & $\bold{59.49}_{\pm 1.19}$ & $75.38_{\pm 0.92}$ & $81.04_{\pm 0.41}$ & $86.03_{\pm 0.09}$ \\
        & StochCA & $\bold{60.14}_{\pm 1.01}$ & $\bold{77.82}_{\pm 0.91}$ & $\bold{83.51}_{\pm 0.74}$ & $\bold{87.55}_{\pm 0.10}$ \\
        \midrule
        \multirow{5}{*}{Car} & FT & $44.94_{\pm 0.18}$ & $\bold{73.73}_{\pm 1.20}$ & $85.20_{\pm 0.24}$ & $91.21_{\pm 0.14}$ \\
        & L2-SP & $44.51_{\pm 0.14}$ & $\bold{73.51}_{\pm 1.16}$ & $85.16_{\pm 0.21}$ & $91.17_{\pm 0.14}$ \\
        & BSS & $44.97_{\pm 0.38}$ & $\bold{73.87}_{\pm 1.27}$ & $85.51_{\pm 0.22}$ & $91.12_{\pm 0.14}$ \\
        & Co-tuning & $44.94_{\pm 0.90}$ & $72.90_{\pm 0.70}$ & $84.62_{\pm 0.41}$ & $90.61_{\pm 0.13}$ \\
        & StochCA & $\bold{48.34}_{\pm 1.22}$ & $\bold{74.68}_{\pm 1.04}$ & $\bold{86.01}_{\pm 0.25}$ & $\bold{91.72}_{\pm 0.20}$ \\
        \midrule
        \multirow{6}{*}{Aircraft} & FT & $55.54_{\pm 1.26}$ & $75.47_{\pm 0.59}$ & $83.06_{\pm 0.35}$ & $\bold{89.04}_{\pm 0.46}$ \\ 
        & L2-SP & $55.12_{\pm 1.29}$ & $75.19_{\pm 0.58}$ & $82.83_{\pm 0.25}$ & $\bold{88.90}_{\pm 0.42}$ \\ 
        & BSS & $58.79_{\pm 1.13}$ & $\bold{77.53}_{\pm 0.82}$ & $\bold{84.20}_{\pm 0.38}$ & $\bold{89.41}_{\pm 0.60}$ \\ 
        & Co-tuning & $56.53_{\pm 1.62}$ & $75.98_{\pm 0.81}$ & $\bold{83.69}_{\pm 0.66}$ & $\bold{89.37}_{\pm 0.39}$ \\ 
        & StochCA & $57.80_{\pm 0.40}$ & $76.40_{\pm 0.44}$ & $\bold{83.65}_{\pm 0.64}$ & $\bold{89.24}_{\pm 0.31}$ \\ 
        \cmidrule{2-6}
        & BSS + StochCA & $\bold{61.30}_{\pm 1.10}$ & $\bold{78.25}_{\pm 0.21}$ & $\bold{84.44}_{\pm 0.33}$ & $\bold{89.49}_{\pm 0.35}$ \\ 
        \midrule
        \multirow{6}{*}{Dog} & FT & $83.87_{\pm 0.23}$ & $88.43_{\pm 0.26}$ & $90.54_{\pm 0.25}$ & $92.54_{\pm 0.26}$ \\ 
        & L2-SP & $84.09_{\pm 0.24}$ & $88.37_{\pm 0.18}$ & $90.74_{\pm 0.17}$ & $92.67_{\pm 0.26}$ \\ 
        & BSS & $83.98_{\pm 0.03}$ & $87.84_{\pm 0.23}$ & $90.36_{\pm 0.24}$ & $92.45_{\pm 0.29}$ \\ 
        & Co-tuning & $89.15_{\pm 0.23}$ & $91.15_{\pm 0.24}$ & $92.53_{\pm 0.16}$ & $93.71_{\pm 0.09}$ \\ 
        & StochCA & $87.75_{\pm 0.30}$ & $90.19_{\pm 0.11}$ & $91.74_{\pm 0.13}$ & $93.18_{\pm 0.10}$ \\ 
        \cmidrule{2-6}
        & Co-Tuning + StochCA & $\bold{90.57}_{\pm 0.30}$ & $\bold{91.88}_{\pm 0.11}$ & $\bold{92.82}_{\pm 0.13}$ & $\bold{93.94}_{\pm 0.10}$ \\ 
        \bottomrule
    \end{tabular}
    }
    \end{center}
    \caption{Transfer learning results with ImageNet pretrained models. \textbf{Bold} font indicates the best performance considering the standard deviation across three random runs. 
}
    \label{tab:tl-img}
\end{table}

\subsection{Experimental Results}
\paragraph{Transfer learning}
Table~\ref{tab:tl-img} presents the performance of StochCA and various transfer learning methods on four benchmarks, utilizing the ImageNet pretrained model. To investigate the effectiveness of transfer learning methods in small data regimes, we experiment with using $15\%$, $30\%$, $50\%$, and $100\%$ of the entire training data, following the previous transfer learning studies~\cite{xuhong2018explicit,chen2019catastrophic,you2020co}. 

As shown in Table~\ref{tab:tl-img}, StochCA achieves the best performance on the CUB and Car datasets across various data sampling rates. Notably, its performance advantage becomes more pronounced with the reduction of training data. This underscores StochCA’s ability to efficiently leverage the pretrained model's knowledge, even with limited training samples. On the other hand, in Aircraft and Dog, StochCA demonstrates the second-best performance. However, the performance can be further improved by combining StochCA with the best-performing methods for each dataset. For example, when combining Co-Tuning and StochCA, the highest performance is achieved across all training data ratios on the Dog dataset. This suggests a synergistic effect when integrating StochCA with existing transfer learning methods. In summary, StochCA demonstrates superior performance in several transfer learning benchmarks. Furthermore, its integration with existing methods can yield even more impressive results, highlighting its complementary benefits in the realm of transfer learning.

\begin{table}[t]
    \begin{center}
    \resizebox{0.8\textwidth}{!}{%
    \begin{tabular}{p{100pt}ccccc}
        \toprule
        \textbf{Method} & \textbf{PACS} & \textbf{VLCS} & \textbf{OfficeHome} & \textbf{DomainNet} & \textbf{Avg.} \\
        \midrule
        \multicolumn{6}{l}{\textit{ImageNet pretrained ViT backbone}} \\
        \midrule
        ERM & $86.9$ & $79.2$ & $73.0$ & $\bold{47.8}$ & $71.7$ \\
        IRM$^{\dagger}$ & $83.9$ & $79.5$ & $72.3$ & $27.1$ & $65.7$ \\
        CDANN$^{\dagger}$ & $82.3$ & $79.1$ & $72.3$ & $34.3$ & $67.0$ \\
        DANN$^{\dagger}$ & $82.6$ & $79.6$ & $71.9$ & $35.0$ & $67.3$ \\
        GDRO$^{\dagger}$ & $86.6$ & $78.9$ & $74.5$ & $41.3$ & $70.3$ \\
        CORAL$^{\dagger}$ & $86.2$ & $79.2$ & $74.5$ & $47.7$ & $71.9$ \\
        MLDG$^{\dagger}$ & $87.0$ & $78.7$ & $\bold{74.8}$ & $47.6$ & $72.0$ \\
        Mixup$^{\dagger}$ & $\bold{87.4}$ & $79.1$ & $74.5$ & $46.7$ & $72.0$ \\
        MMD$^{\dagger}$ & $86.9$ & $79.8$ & $74.5$ & $47.4$ & $72.2$ \\
        \textcolor{lightgray}{DoPrompt$^{\dagger}$} & \color{lightgray}$88.1$ & \color{lightgray}$80.4$ & \color{lightgray}$76.0$ & \color{lightgray}$48.3$ & \color{lightgray}$73.2$ \\
        DoPrompt & $86.7$ & $78.9$ & $72.8$ & $47.4$ & $71.5$ \\
        StochCA & $\bold{87.4}$ & $\bold{80.0}$ & $74.6$ & $47.7$ & $\bold{72.4}$ \\
        \midrule
        \multicolumn{6}{l}{\textit{CLIP(400M) pretrained ViT backbone}} \\
        \midrule
        ERM$^{\ddagger}$ & $83.4$ & $75.9$ & $66.4$ & $44.4$ & $67.5$ \\
        MIRO$^{\ddagger}$ & $95.6$ & $82.2$ & $82.5$ & $54.0$ & $78.6$ \\
        StochCA & $94.8$ & $82.1$ & $82.2$ & $58.0$ & $79.3$ \\
        MIRO+StochCA & $\bold{96.4}$ & $\bold{83.0}$ & $\bold{85.1}$ & $\bold{60.0}$ & $\bold{81.1}$ \\
        \bottomrule
    \end{tabular}
    }
    \end{center}
    \caption{Domain generalization performance using ImageNet and CLIP ViT backbones. The best performance for each dataset is highlighted in \textbf{bold}. Performances marked with $^{\dagger}$ and $^{\ddagger}$ refer to the results reported in~\cite{zheng2022prompt} and~\cite{cha2022miro}, respectively.}
    \label{tab:dg}
\end{table}

\paragraph{Domain generalization}
To further investigate the robustness of StochCA, we conduct experiments within domain generalization settings. The overall comparison results are summarized in Table~\ref{tab:dg}, which reports the average test accuracy for all source-target domain combinations for each dataset. While most results are directly obtained from previous studies~\cite{zheng2022prompt,cha2022miro}, we encountered challenges in reproducing the performance of DoPrompt~\cite{zheng2022prompt} using the author's official code. To maintain fairness, the reported performance of DoPrompt is noted in gray font, with our reproduced results provided for comparison. 

As shown in Table~\ref{tab:dg}, StochCA demonstrates competitive performance across all benchmarks with the ImageNet pretrained model. Notably, on the PACS dataset, StochCA achieves a top performance of 87.4\%, matching the highest result attained by Mixup. Overall, it is observed that StochCA demonstrates the highest average performance across all benchmark datasets. Furthermore, we extended our experiments to include the CLIP pretrained model~\cite{radford2021learning} as a backbone, which is trained on a significantly larger-scale dataset compared to ImageNet (i.e., 400M vs. 1.3M of training data). As shown in the lower section of Table~\ref{tab:dg}, StochCA outperforms all previous methods by a considerable margin when employing a larger-scale backbone. Remarkably, the average performance of StochCA exceeds ERM by $11.8\%$ and even surpasses MIRO~\cite{cha2022miro}, the current state-of-the-art method in DG, by $0.7\%$. These findings highlight the effectiveness of selectively leveraging information through cross-attention with a pretrained model. Additionally, integrating StochCA with MIRO consistently yields the best results across all datasets, demonstrating substantial performance improvements. For example, the average performance improvement of MIRO+StochCA over ERM is $13.6\%$. This observation further reinforces the complementary potential of the proposed method when used in conjunction with other approaches. Detailed results for each dataset are provided in Table~\ref{tab:dg_pa_vl} and Table~\ref{tab:dg_oh_dn}.

\begin{table}[t]
    \centering
    \resizebox{0.465\textwidth}{!}{%
    \begin{tabular}{p{50pt}ccccc}
        \toprule
        \textbf{Method} & \textbf{Art} & \textbf{Cartoon} & \textbf{Photo} & \textbf{Sketch} & \textbf{Avg.} \\
        \midrule
        ERM & $90.3_{\pm0.8}$ & $84.2_{\pm0.5}$ & $99.3_{\pm0.2}$ & $73.9_{\pm3.2}$ & $86.9$ \\
        IRM$^{\dagger}$ & $89.1_{\pm0.7}$ & $79.1_{\pm1.1}$ & $98.3_{\pm0.1}$ & $68.2_{\pm5.1}$ & $83.9$ \\
        CDANN$^{\dagger}$ & $87.3_{\pm0.6}$ & $80.1_{\pm0.5}$ & $98.5_{\pm0.2}$ & $63.1_{\pm2.9}$ & $82.3$ \\
        DANN$^{\dagger}$ & $87.5_{\pm0.4}$ & $78.6_{\pm0.9}$ & $98.0_{\pm0.2}$ & $66.3_{\pm1.7}$ & $82.6$ \\
        GDRO$^{\dagger}$ & $90.1_{\pm0.6}$ & $80.5_{\pm1.0}$ & $98.8_{\pm0.1}$ & $77.1_{\pm1.5}$ & $86.6$ \\
        CORAL$^{\dagger}$ & $89.7_{\pm0.3}$ & $80.2_{\pm1.0}$ & $98.7_{\pm0.2}$ & $76.1_{\pm1.6}$ & $86.2$ \\
        MLDG$^{\dagger}$ & $90.4_{\pm0.5}$ & $83.0_{\pm0.7}$ & $98.7_{\pm0.0}$ & $76.0_{\pm0.2}$ & $87.0$ \\
        Mixup$^{\dagger}$ & $91.1_{\pm0.1}$ & $84.0_{\pm0.3}$ & $99.2_{\pm0.2}$ & $75.4_{\pm0.5}$ & $87.4$ \\
        MMD$^{\dagger}$ & $89.9_{\pm0.5}$ & $80.8_{\pm0.5}$ & $98.7_{\pm0.1}$ & $77.1_{\pm0.7}$ & $86.6$ \\
        \textcolor{lightgray}{DoPrompt$^{\dagger}$} & \color{lightgray}$91.1_{\pm0.3}$ & \color{lightgray}$83.0_{\pm0.2}$ & \color{lightgray}$99.6_{\pm0.0}$ & \color{lightgray}$78.7_{\pm0.8}$ & \color{lightgray}$88.1$ \\
        DoPrompt & $89.9_{\pm2.2}$ & $84.1_{\pm0.7}$ & $98.9_{\pm0.0}$ & $74.1_{\pm3.3}$ & $86.7$ \\
        StochCA & $90.8_{\pm0.6}$ & $82.9_{\pm0.7}$ & $98.9_{\pm0.1}$ & $77.1_{\pm2.4}$ & $87.4$ \\
        \bottomrule
    \end{tabular}
    }
    \resizebox{0.5\textwidth}{!}{%
    \begin{tabular}{p{50pt}ccccc}
        \toprule
        \textbf{Method} & \textbf{Caltech101} & \textbf{LabelMe} & \textbf{SUN09} & \textbf{VOC2007} & \textbf{Avg.} \\
        \midrule
        ERM & $96.2_{\pm0.5}$ & $64.4_{\pm0.9}$ & $77.3_{\pm0.3}$ & $79.0_{\pm1.0}$ & $79.2$ \\
        IRM$^{\dagger}$ & $96.2_{\pm0.5}$ & $65.7_{\pm0.8}$ & $76.8_{\pm0.4}$ & $77.7_{\pm0.9}$ & $79.5$ \\
        CDANN$^{\dagger}$ & $96.3_{\pm0.2}$ & $66.3_{\pm0.5}$ & $76.5_{\pm0.6}$ & $77.3_{\pm0.5}$ & $79.1$ \\
        DANN$^{\dagger}$ & $97.7_{\pm0.3}$ & $66.5_{\pm0.5}$ & $76.0_{\pm1.0}$ & $78.0_{\pm0.7}$ & $79.6$ \\
        GDRO$^{\dagger}$ & $97.1_{\pm0.5}$ & $65.2_{\pm0.7}$ & $75.1_{\pm1.0}$ & $78.3_{\pm1.0}$ & $78.9$ \\
        CORAL$^{\dagger}$ & $96.2_{\pm0.7}$ & $66.0_{\pm0.3}$ & $76.8_{\pm0.6}$ & $78.0_{\pm0.9}$ & $79.2$ \\
        MLDG$^{\dagger}$ & $95.7_{\pm0.5}$ & $65.9_{\pm0.3}$ & $76.3_{\pm0.8}$ & $78.3_{\pm1.0}$ & $78.9$ \\
        Mixup$^{\dagger}$ & $97.1_{\pm0.9}$ & $65.4_{\pm0.2}$ & $76.1_{\pm0.4}$ & $78.0_{\pm0.3}$ & $79.1$ \\
        MMD$^{\dagger}$ & $96.8_{\pm0.1}$ & $65.6_{\pm1.0}$ & $77.3_{\pm0.6}$ & $79.3_{\pm1.5}$ & $79.8$ \\
        \textcolor{lightgray}{DoPrompt$^{\dagger}$} & \color{lightgray}$97.2_{\pm0.1}$ & \color{lightgray}$67.4_{\pm0.5}$ & \color{lightgray}$77.4_{\pm1.0}$ & \color{lightgray}$79.7_{\pm0.5}$ & \color{lightgray}$80.4$ \\
        DoPrompt & $96.2_{\pm0.6}$ & $64.5_{\pm0.8}$ & $76.8_{\pm1.5}$ & $78.0_{\pm1.1}$ & $78.9$ \\
        StochCA & $96.3_{\pm0.3}$ & $65.9_{\pm0.8}$ & $77.9_{\pm0.6}$ & $79.9_{\pm0.7}$ & $80.0$ \\
        \bottomrule
    \end{tabular}
    }
    \caption{Detailed results on PACS (left) and VLCS (right). Performances marked with $^{\dagger}$ refer to the results reported in~\cite{zheng2022prompt}.}
    \label{tab:dg_pa_vl}
\end{table}

\begin{table}[!t]
    \centering
    \resizebox{0.4\textwidth}{!}{%
    \begin{tabular}{p{50pt}ccccc}
        \toprule
        \textbf{Method} & \textbf{Art} & \textbf{Clipart} & \textbf{Product} & \textbf{Real} & \textbf{Avg.} \\
        \midrule
        ERM & $69.8_{\pm0.5}$ & $59.4_{\pm1.0}$ & $79.6_{\pm0.5}$ & $83.1_{\pm0.4}$ & $73.0$ \\
        IRM$^{\dagger}$ & $69.5_{\pm0.1}$ & $57.1_{\pm0.3}$ & $79.1_{\pm0.1}$ & $83.3_{\pm0.1}$ & $72.3$ \\
        CDANN$^{\dagger}$ & $68.2_{\pm0.4}$ & $56.8_{\pm0.4}$ & $79.8_{\pm0.4}$ & $84.2_{\pm0.0}$ & $72.3$ \\
        DANN$^{\dagger}$ & $68.7_{\pm0.9}$ & $55.8_{\pm0.4}$ & $79.2_{\pm0.0}$ & $83.9_{\pm0.2}$ & $71.9$ \\
        GDRO$^{\dagger}$ & $71.2_{\pm0.7}$ & $60.9_{\pm0.4}$ & $81.1_{\pm0.3}$ & $84.7_{\pm0.3}$ & $74.5$ \\
        CORAL$^{\dagger}$ & $72.3_{\pm0.7}$ & $60.9_{\pm1.0}$ & $80.4_{\pm0.4}$ & $84.3_{\pm0.0}$ & $74.5$ \\
        MLDG$^{\dagger}$ & $72.3_{\pm0.0}$ & $60.0_{\pm1.1}$ & $82.2_{\pm0.2}$ & $84.7_{\pm0.1}$ & $74.8$ \\
        Mixup$^{\dagger}$ & $72.7_{\pm0.4}$ & $59.7_{\pm0.5}$ & $80.9_{\pm0.3}$ & $84.6_{\pm0.2}$ & $74.5$ \\
        MMD$^{\dagger}$ & $71.8_{\pm0.2}$ & $60.7_{\pm1.0}$ & $81.4_{\pm0.3}$ & $84.3_{\pm0.1}$ & $74.5$ \\
        \textcolor{lightgray}{DoPrompt$^{\dagger}$} & \color{lightgray}$72.7_{\pm0.5}$ & \color{lightgray}$62.3_{\pm0.6}$ & \color{lightgray}$83.2_{\pm0.1}$ & \color{lightgray}$85.9_{\pm0.1}$ & \color{lightgray}$76.0$ \\
        DoPrompt & $69.5_{\pm0.5}$ & $59.1_{\pm0.3}$ & $79.9_{\pm0.5}$ & $82.7_{\pm0.2}$ & $72.8$ \\
        StochCA & $72.4_{\pm0.8}$ & $60.1_{\pm0.5}$ & $81.6_{\pm0.2}$ & $84.3_{\pm0.2}$ & $74.6$ \\
        \bottomrule
    \end{tabular}
    }
    \resizebox{0.565\textwidth}{!}{%
    \begin{tabular}{p{50pt}ccccccc}
        \toprule
        \textbf{Method} & \textbf{Clipart} & \textbf{Infograph} & \textbf{Painting} & \textbf{Quickdraw} & \textbf{Real} & \textbf{Sketch} & \textbf{Avg.} \\
        \midrule
        ERM & $67.6_{\pm0.1}$ & $23.9_{\pm0.3}$ & $54.2_{\pm0.4}$ & $16.7_{\pm0.3}$ & $69.1_{\pm0.0}$ & $54.9_{\pm0.2}$ & $47.8$ \\
        IRM$^{\dagger}$ & $35.6_{\pm0.5}$ & $13.9_{\pm0.2}$ & $33.1_{\pm0.8}$ & $6.4_{\pm0.2}$ & $40.6_{\pm0.4}$ & $32.8_{\pm0.3}$ & $27.1$ \\
        CDANN$^{\dagger}$ & $45.3_{\pm0.5}$ & $13.5_{\pm0.0}$ & $45.8_{\pm0.0}$ & $7.7_{\pm0.2}$ & $49.6_{\pm0.1}$ & $43.8_{\pm0.2}$ & $34.3$ \\
        DANN$^{\dagger}$ & $45.6_{\pm0.1}$ & $14.4_{\pm0.2}$ & $44.6_{\pm0.6}$ & $8.1_{\pm0.0}$ & $52.3_{\pm0.3}$ & $44.8_{\pm0.5}$ & $35.0$ \\
        GDRO$^{\dagger}$ & $59.3_{\pm0.2}$ & $20.8_{\pm0.2}$ & $46.6_{\pm0.6}$ & $11.3_{\pm0.5}$ & $63.7_{\pm0.1}$ & $47.0_{\pm0.3}$ & $41.3$ \\
        CORAL$^{\dagger}$ & $66.8_{\pm0.2}$ & $24.4_{\pm0.1}$ & $54.6_{\pm0.2}$ & $16.2_{\pm0.2}$ & $69.3_{\pm0.1}$ & $54.9_{\pm0.3}$ & $47.7$ \\
        MLDG$^{\dagger}$ & $65.8_{\pm1.4}$ & $23.8_{\pm0.1}$ & $54.7_{\pm0.1}$ & $17.2_{\pm0.2}$ & $69.0_{\pm0.1}$ & $54.9_{\pm0.3}$ & $47.6$ \\
        Mixup$^{\dagger}$ & $65.1_{\pm0.1}$ & $23.7_{\pm0.1}$ & $54.4_{\pm0.1}$ & $15.1_{\pm0.3}$ & $67.9_{\pm0.0}$ & $54.0_{\pm0.1}$ & $46.7$ \\
        MMD$^{\dagger}$ & $67.0_{\pm0.2}$ & $23.8_{\pm0.2}$ & $54.0_{\pm0.0}$ & $15.9_{\pm0.5}$ & $69.0_{\pm0.1}$ & $54.6_{\pm0.2}$ & $47.4$ \\
        \textcolor{lightgray}{DoPrompt$^{\dagger}$} & \color{lightgray}$67.7_{\pm0.2}$ & \color{lightgray}$24.6_{\pm0.1}$ & \color{lightgray}$54.9_{\pm0.1}$ & \color{lightgray}$17.5_{\pm0.2}$ & \color{lightgray}$69.6_{\pm0.3}$ & \color{lightgray}$55.2_{\pm0.5}$ & \color{lightgray}$48.3$ \\
        DoPrompt & $66.6_{\pm0.3}$ & $23.4_{\pm0.2}$ & $54.5_{\pm0.2}$ & $16.2_{\pm0.2}$ & $69.2_{\pm0.3}$ & $54.5_{\pm0.0}$ & $47.4$ \\
        StochCA & $67.0_{\pm0.3}$ & $23.9_{\pm0.9}$ & $54.5_{\pm0.3}$ & $16.8_{\pm0.3}$ & $69.0_{\pm0.2}$ & $54.7_{\pm0.4}$ & $47.7$  \\
        \bottomrule
    \end{tabular}
    }
    \caption{Detailed results on OfficeHome (left) and DomainNet (right). Performances marked with $^{\dagger}$ refer to the results reported in~\cite{zheng2022prompt}.}
    \label{tab:dg_oh_dn}
\end{table}

\paragraph{Impact of StochCA on query, key, and value}
In this section, we delve into the impact of StochCA. We hypothesize that if StochCA effectively references the knowledge of the pretrained model through cross-attention, then the query, key, and value from the target model should exhibit similarities to those from the pretrained model. Alternatively, one might explicitly incorporate regularization techniques, such as $L_2$ penalization (e.g., minimizing the $L_2$ norm of the discrepancy between keys from the pretrained and target models), to guarantee this similarity. We denote this method as L2-Reg. To test our hypothesis, we investigate FT (i.e., vanilla fine-tuning), StochCA, and L2-Reg by assessing the cosine similarity between the queries, keys, and values from the fine-tuned target model and those from the pretrained model. 

Table~\ref{tab:cos} shows the cosine similarities calculated at each attention layer for the CUB and OfficeHome datasets. We use $15\%$ of the training data for the CUB dataset, and adopt source-target domain setting of \{Clipart, Product, Real\} $\rightarrow{}$ Art for the OfficeHome dataset. The results reveal that StochCA consistently shows higher cosine similarity compared to FT (vanilla fine-tuning), suggesting that it facilitates learning features akin to those of the pretrained model during fine-tuning. L2-Reg, which applies direct regularization, demonstrates even higher similarity than StochCA. However, in terms of test performance, StochCA outperforms both L2-Reg and FT, achieving the highest scores: $60.1\% $ (StochCA) $> 58.1\% $ (L2-Reg) $> 56.1\%$ (FT) in CUB and $74.6\% $ (StochCA) $> 73.7\% $ (L2-Reg) $> 73.0\%$ (FT) in OfficeHome. These findings suggest that StochCA not only allows the selective retrieval of useful information from the pretrained knowledge but also mitigates the risk of over-reliance on the pretrained model, which could lead to negative transfer. This balance is key to its superior performance compared to alternative approaches.

\begin{table}[t]
    \begin{center}
    \resizebox{0.75\textwidth}{!}{%
    \begin{tabular}{lccccccccc}
        \toprule
        \multirow{2}{*}{Layer} & \multicolumn{3}{c}{Query} & \multicolumn{3}{c}{Key} & \multicolumn{3}{c}{Value}  \\
        \cmidrule{2-10}
        & FT & StochCA & L2-Reg & FT & StochCA & L2-Reg & FT & StochCA & L2-Reg \\
        \midrule
        $1$ & $0.98$ & $0.99$ & $1.00$ & $0.99$ & $1.00$ & $1.00$ & $0.86$ & $0.97$ & $0.99$ \\
        $2$ & $0.97$ & $0.99$ & $1.00$ & $0.94$ & $0.99$ & $1.00$ & $0.85$ & $0.95$ & $0.99$ \\
        $3$ & $0.97$ & $0.99$ & $1.00$ & $0.96$ & $0.99$ & $1.00$ & $0.90$ & $0.97$ & $1.00$ \\
        $4$ & $0.97$ & $0.99$ & $1.00$ & $0.96$ & $0.98$ & $1.00$ & $0.92$ & $0.97$ & $1.00$ \\
        $5$ & $0.95$ & $0.98$ & $1.00$ & $0.95$ & $0.98$ & $1.00$ & $0.88$ & $0.96$ & $1.00$ \\
        $6$ & $0.91$ & $0.96$ & $1.00$ & $0.91$ & $0.96$ & $1.00$ & $0.84$ & $0.94$ & $0.99$ \\
        $7$ & $0.89$ & $0.95$ & $1.00$ & $0.88$ & $0.95$ & $1.00$ & $0.80$ & $0.92$ & $0.99$ \\
        $8$ & $0.85$ & $0.93$ & $1.00$ & $0.84$ & $0.93$ & $0.99$ & $0.77$ & $0.90$ & $0.99$ \\
        $9$ & $0.83$ & $0.91$ & $0.99$ & $0.84$ & $0.92$ & $0.99$ & $0.71$ & $0.87$ & $0.98$ \\
        $10$ & $0.88$ & $0.94$ & $1.00$ & $0.85$ & $0.93$ & $0.99$ & $0.66$ & $0.85$ & $0.97$ \\
        $11$ & $0.86$ & $0.94$ & $0.99$ & $0.85$ & $0.93$ & $0.99$ & $0.66$ & $0.85$ & $0.97$ \\
        $12$ & $0.74$ & $0.89$ & $0.98$ & $0.86$ & $0.93$ & $0.99$ & $0.60$ & $0.84$ & $0.95$ \\
        \midrule
        Avg. & $0.90$ & $0.96$ & $1.00$ & $0.90$ & $0.96$ & $1.00$ & $0.79$ & $0.92$ & $0.98$ \\
        \bottomrule
    \end{tabular}
    }
    \end{center}
    \begin{center}
    \resizebox{0.75\textwidth}{!}{%
    \begin{tabular}{lccccccccc}
        \toprule
        \multirow{2}{*}{Layer} & \multicolumn{3}{c}{Query} & \multicolumn{3}{c}{Key} & \multicolumn{3}{c}{Value}  \\
        \cmidrule{2-10}
        & FT & StochCA & L2-Reg & FT & StochCA & L2-Reg & FT & StochCA & L2-Reg \\
        \midrule
        $1$ & $0.99$ & $0.99$ & $0.99$ & $0.99$ & $1.00$ & $0.98$ & $0.91$ & $0.92$ & $0.84$ \\
        $2$ & $0.97$ & $0.98$ & $0.98$ & $0.95$ & $0.96$ & $0.98$ & $0.90$ & $0.91$ & $0.90$ \\
        $3$ & $0.98$ & $0.98$ & $0.99$ & $0.97$ & $0.98$ & $0.98$ & $0.92$ & $0.94$ & $0.90$ \\
        $4$ & $0.97$ & $0.98$ & $0.99$ & $0.97$ & $0.97$ & $0.99$ & $0.94$ & $0.95$ & $0.96$ \\
        $5$ & $0.96$ & $0.97$ & $0.98$ & $0.96$ & $0.97$ & $0.99$ & $0.91$ & $0.93$ & $0.96$ \\
        $6$ & $0.93$ & $0.94$ & $0.98$ & $0.92$ & $0.93$ & $0.98$ & $0.87$ & $0.89$ & $0.95$ \\
        $7$ & $0.89$ & $0.91$ & $0.97$ & $0.89$ & $0.91$ & $0.97$ & $0.83$ & $0.86$ & $0.94$ \\
        $8$ & $0.85$ & $0.88$ & $0.96$ & $0.83$ & $0.87$ & $0.95$ & $0.81$ & $0.85$ & $0.93$ \\
        $9$ & $0.83$ & $0.87$ & $0.95$ & $0.82$ & $0.86$ & $0.95$ & $0.75$ & $0.80$ & $0.90$ \\
        $10$ & $0.85$ & $0.83$ & $0.96$ & $0.83$ & $0.87$ & $0.94$ & $0.72$ & $0.77$ & $0.87$ \\
        $11$ & $0.73$ & $0.87$ & $0.96$ & $0.79$ & $0.83$ & $0.93$ & $0.69$ & $0.76$ & $0.86$ \\
        $12$ & $0.90$ & $0.80$ & $0.94$ & $0.85$ & $0.89$ & $0.95$ & $0.70$ & $0.76$ & $0.87$ \\
        \midrule
        Avg. & $0.90$ & $0.92$ & $0.97$ & $0.90$ & $0.92$ & $0.97$ & $0.83$ & $0.86$ & $0.90$ \\
        \bottomrule
    \end{tabular}
    }
    \end{center}
    \caption{Cosine similarity comparison for the query, key, and value vectors between the fine-tuned target model (through FT, StochCA, and L2-Reg) and the pretrained model across all layers: (Upper) the CUB dataset and (Lower) the OfficeHome dataset.}
    \label{tab:cos}
\end{table}

\subsection{Ablation Study}
In this section, we perform an ablation study to investigate the impact of the StochCA components in two aspects: the cross-attention probability $p$ and the specific design of stochastic cross-attention. Table~\ref{tab:abl_caprob} presents a comparative analysis of transfer learning performance with varying values of $p$.
Our findings indicate that, except for the Dog dataset, performance generally declines with increasing $p$. This trend suggests that excessive reliance on the pretrained model may lead to suboptimal performance. 
For example, in the Aircraft dataset, StochCA with $p=0.1$ surpasses FT, but the performance diminishes for $p$ values exceeding $0.5$. In contrast, the Dog dataset shows a slight improvement in performance as $p$ increases. This can be attributed to the extensive representation of dog images in ImageNet, making a higher reliance on the pretrained model more beneficial in this context. 
In summary, the optimal probability of cross-attention, which determines the extent of knowledge reference from the pretrained model, varies across datasets. When there is a significant semantic similarity between the pretraining and target datasets, a higher cross-attention probability can lead to enhanced performance.

\begin{table}[t]
    \begin{center}
    \resizebox{0.7\textwidth}{!}{%
    \begin{tabular}{c|cccccccc}
        \toprule
        \multirow{2}{*}{CA prob} & \multicolumn{2}{c}{CUB} & \multicolumn{2}{c}{Car} & \multicolumn{2}{c}{Aircraft} & \multicolumn{2}{c}{Dog} \\
        \cmidrule{2-9}
        & $15\%$ & $100\%$ & $15\%$ & $100\%$ & $15\%$ & $100\%$ & $15\%$ & $100\%$  \\
        \midrule 
        No ($=$ FT) & $56.09$ & $86.85$ & $44.94$ & $91.21$ & $55.54$ & $89.04$ & $83.87$ & $92.54$  \\
        \midrule 
        0.1 & $\bold{\underline{60.14}}$ & $\bold{\underline{87.55}}$ & $\bold{\underline{48.34}}$ & $\bold{\underline{91.72}}$ & $\bold{\underline{57.80}}$ & $\bold{\underline{89.24}}$ & $87.05$ & $92.94$ \\
        0.3 & $59.69$ & $87.56$ & $47.27$ & $91.00$ & $56.10$ & $88.73$ & $87.63$ & $93.12$ \\
        0.5 & $58.19$ & $87.22$ & $44.75$ & $89.80$ & $52.53$ & $87.34$ & $\underline{87.75}$ & $\bold{\underline{93.18}}$ \\
        0.7 & $56.62$ & $86.50$ & $41.94$ & $87.70$ & $49.42$ & $85.38$ & $\bold{88.01}$ & $92.93$ \\
        \bottomrule
    \end{tabular}
    }
    \end{center}
    \caption{Ablation experiments on the cross-attention probability $p$ of StochCA. The highest performance in each case is highlighted in \textbf{bold}, while \underline{underlined} values indicate the cross-attention probability chosen based on validation accuracy.}
    \label{tab:abl_caprob}
\end{table}

\begin{table}[t]
    \begin{center}
    \resizebox{0.5\textwidth}{!}{%
    \begin{tabular}{lcccc}
        \toprule
        \multirow{2}{*}{Method} & \multicolumn{2}{c}{CUB} & \multicolumn{2}{c}{Car} \\
        \cmidrule{2-5}
        & $15\%$ & $100\%$ & $15\%$ & $100\%$ \\
        \midrule 
        FT & $56.09$ & $86.85$ & $44.94$ & $91.21$ \\
        \midrule 
        FT+CA & $58.84$ & $87.48$ & 44.71 & $90.51$ \\
        FT+CA (only SA) & $58.62$ & $87.18$ & $45.91$ & $91.43$  \\
        StochCA & $\bold{60.14}$ & $\bold{87.55}$ & $\bold{48.34}$ & $\bold{91.72}$ \\
        \bottomrule
    \end{tabular}
    }
    \end{center}
    \caption{Comparison between StochCA and alternative combinations of self-attention and cross-attention}
    \label{tab:abl_saca}
\end{table}

Next, we evaluate various design choices for integrating self-attention (SA) and cross-attention (CA).  Table~\ref{tab:abl_saca} presents the comparative results between StochCA, which implements block-wise cross-attention stochastically, and other methods maintaining separate paths for SA and CA. The method FT+CA represents an ensemble approach where the outputs from both SA and CA paths are averaged during inference after training, as depicted in Figure~\ref{fig:simple} (left). Since this FT+CA method requires additional memory and computational costs due to the necessity of maintaining the pretrained model during inference, we also assess an alternative approach, FT+CA (only SA), where the learning process is identical but only the output of the SA path is utilized for inference. 
As indicated in Table~\ref{tab:abl_saca}, both alternative approaches, FT+CA and FT+CA (only SA), are superior to FT in most cases, although the amount of improvement remains modest. In contrast, StochCA demonstrates significant performance gains over these simpler SA and CA combinations. These results suggest that in addition to its computational efficiency, StochCA has a regularization effect due to its stochastic nature (i.e., block-wise selection between SA and CA), leading to superior performance. 

\section{Conclusion}
\label{sec:Con}
In the current trend of scaling up pretraining, the effective utilization of pretrained models with large-scale datasets becomes increasingly important. In this study, we propose a novel fine-tuning method called StochCA, which selectively leverages useful information from pretrained models that possess extensive knowledge learned from large-scale datasets. Specifically, cross-attention is introduced as a mechanism for referencing pretrained knowledge that is useful for a target task, involving the interaction between queries from the target model and keys and values from the pretrained model. In addition, by stochastically implementing either cross-attention or self-attention in a block-wise manner, StochCA avoids incurring additional computational costs while achieving a substantial regularization effect. Our experimental results on transfer learning and domain generalization show the superiority of the proposed method. Moreover, it is noteworthy that StochCA demonstrates complementary effects when integrated with existing methods in both tasks. 

While our proposed method has been demonstrated on image classification tasks, its applicability is not limited to this task. StochCA is adaptable to various computer vision tasks, provided that the underlying pretrained model utilizes the ViT architecture. To broaden its applicability, it would be valuable to verify StochCA on object detection or semantic segmentation tasks, which we leave for future work. Additionally, exploring the synergy between StochCA and knowledge distillation strategies offers an intriguing prospect for advancing research in this area.

\section*{Acknowledgement}
\noindent This work was supported by the National Research Foundation of Korea (NRF) grant funded by the Korea government (MSIT and the Ministry of Education) (NRF-2021R1C1C1011907 and NRF-2019R1A6A1A03032119).

\bibliographystyle{unsrt}  
\bibliography{references}

\end{document}